\documentclass[10pt,twocolumn,letterpaper]{article}

\usepackage[11]{wacv} 

\usepackage{booktabs,multirow,makecell,tabularx}
\usepackage{pifont}            
\usepackage{amsmath,amssymb}
\newcommand{\cmark}{{\ding{51}}}
\newcommand{\xmark}{{\ding{55}}}
\newcommand{\best}[1]{\textbf{#1}}      
\newcommand{\second}[1]{\underline{#1}} 
\newcommand{\SRCC}{SRCC}
\newcommand{\RLtwo}{$\text{R-}\ell_{2}$}
\usepackage{graphicx}
\usepackage{xcolor} 

\newcommand{\mypara}[1]{%
  \vspace{0.2em}\noindent\textbf{#1}\quad
}

\definecolor{wacvblue}{rgb}{0.21,0.49,0.74}
\usepackage[pagebackref,breaklinks,colorlinks,allcolors=wacvblue]{hyperref}

\def\wacvPaperID{2024} 
\def\confName{WACV}
\def\confYear{2026}

\title{
CaFlow: Enhancing Long-Term Action Quality Assessment with Causal Counterfactual Flow
}

\author{
Ruisheng Han$^1$, Kanglei Zhou$^2$, Shuang Chen$^1$, Amir Atapour-Abarghouei$^1$, Hubert P. H. Shum$^1$\thanks{Corresponding author. This research is supported in part by the EPSRC NortHFutures project (ref: EP/X031012/1).} \\
$^1$Durham University \quad $^2$Tsinghua University\\ 
{\tt\small \{ruisheng.han, shuang.chen, amir.atapour-abarghouei, hubert.shum\}@durham.ac.uk,}\\
{\tt\small \{zhoukanglei\}@tsinghua.edu.cn}
}

\begin{document}
\maketitle
\begin{abstract}

Action Quality Assessment (AQA) predicts fine-grained execution scores from action videos and is widely applied in sports, rehabilitation, and skill evaluation. \textit{Long-term AQA}, as in figure skating or rhythmic gymnastics, is especially challenging since it requires modeling extended temporal dynamics while remaining robust to contextual confounders. Existing approaches either depend on costly annotations or rely on unidirectional temporal modeling, making them vulnerable to spurious correlations and unstable long-term representations. To this end, we propose \textbf{CaFlow}, a unified framework that integrates counterfactual de-confounding with bidirectional time-conditioned flow. The \textit{Causal  Counterfactual Regularization (CCR)} module disentangles causal and confounding features in a self-supervised manner and enforces causal robustness through counterfactual interventions, while the \textit{BiT-Flow} module models forward and backward dynamics with a cycle-consistency constraint to produce smoother and more coherent representations. Extensive experiments on multiple long-term AQA benchmarks demonstrate that CaFlow achieves state-of-the-art performance. Code is available at \url{https://github.com/Harrison21/CaFlow}
\end{abstract}
    
\section{Introduction}

\begin{figure*}[t]
    \centering
    \includegraphics[width=\textwidth]{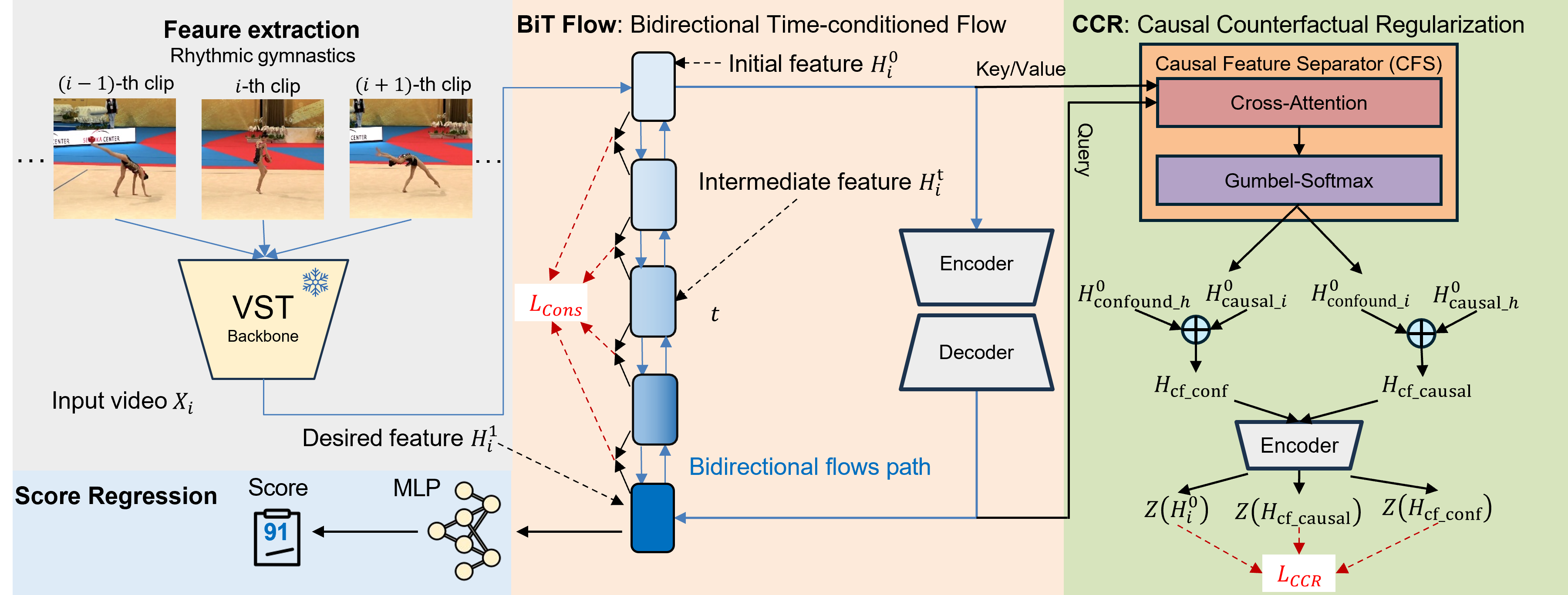}
    \caption{Framework of CaFlow. Our method tackles confounding and domain shift in AQA with two key contributions: (1) Causal Counterfactual Regularization (CCR), which uses a Causal Feature Separator and counterfactual mixing to separate causal from confounding clips and impose a triplet‑style causal loss; (2) Bidirectional Time-conditioned Flow (BiT Flow), a time‑conditioned bidirectional flow that progressively transforms \(H_i^{0}\) to the AQA‑specific representation \(H_i^{1}\) with forward-backward consistency and optimal‑transport regularization. The refined representation is finally regressed by an MLP to the quality score.}
    \label{fig:overview}
    \vspace{-0.8em}
\end{figure*}

Action Quality Assessment (AQA) \cite{zhou2024magr,ji2023localization,zhang2023logo,xu2024fineparser,zhou2025continual} seeks to evaluate how well an action is performed in video sequences, going beyond action recognition to measure fine-grained execution quality \cite{zhou2023hierarchical,zhou2024comprehensive}. Accurate AQA is crucial for applications such as sports analytics \cite{li2022pairwise,yu2021group,zhou2023hierarchical,yao2023contrastive}, medical rehabilitation \cite{zhou2023video,deb2022graph}, and skill assessment \cite{ingwersen2023video,trinh2023self,ding2023sedskill}. Long-term AQA \cite{zeng2020hybrid,xu2022likert}, which evaluates extended video sequences instead of short clips, provides a more comprehensive and practical measure of performance but also introduces greater modeling challenges. 

Recent approaches have attempted to address long-term AQA through representation learning, temporal modeling, or causal reasoning. Likert scoring \cite{xu2022likert} and hierarchical graph methods \cite{zhou2023hierarchical} capture temporal dependencies but struggle with robustness under domain shift. Causal AQA methods such as FineCausal \cite{han2025finecausal} reduce spurious correlations by balancing foreground and background features but rely on costly human-annotated masks, limiting scalability. Meanwhile, advanced refinement approaches like PHI \cite{zhou2025phi} model temporal dynamics but remain restricted to unidirectional flows, which accumulate errors over time and lead to unstable long-term representations.

Despite recent progress, we identify two fundamental yet overlooked challenges in existing methods: (i) \textbf{Confounding bias}, where irrelevant contextual factors such as background, environment, or camera viewpoint are entangled with quality scores, leading to spurious correlations. (ii) \textbf{Temporal instability}, where unidirectional refinement produces noisy or inconsistent features over long sequences, preventing accurate modeling of execution dynamics. Addressing both challenges simultaneously is essential for robust, interpretable, and generalizable long-term AQA. 

To this end, we propose \textbf{CaFlow}, a unified framework that integrates two innovations: a mask-free Causal Counterfactual Regularization (CCR), which leverages the transformer’s ``desired feature'' to partition features and enforce causal robustness via counterfactual testing; and Bidirectional Time-conditioned Flow (BiT-Flow), which refines features by modeling forward-backward dynamics with cycle consistency to yield smoother temporal trajectories. Together, these designs ensure that CaFlow focuses on causal cues while producing stable, temporally coherent features for long-term AQA. We validate CaFlow on three long-term AQA benchmarks: RG (rhythmic gymnastics, four apparatus), FIS-V (figure skating, two scores), and LOGO (diving, single score), totaling over 4,000 videos. Across all datasets, CaFlow achieves state-of-the-art performance. Our \textbf{source code} can be found in the supplementary material. Our main contributions are as follows:
\begin{itemize}[noitemsep,topsep=0pt,partopsep=0pt,parsep=0pt,leftmargin=*]
    \item We introduce a self-supervised causal regularization module (CCR) that separates causal and confounding features without external annotations.
    \item We propose a bidirectional temporal refinement module (BiT-Flow) that enforces forward-backward consistency for stable long-term modeling.
    \item We present a unified causal-temporal framework (CaFlow) that achieves state-of-the-art performance on three long-term AQA benchmarks (RG, Fis-V, LOGO).
\end{itemize}
\vspace{-1em}
\section{Related Work}
\subsection{Action Quality Assessment}
Research on Action Quality Assessment (AQA) has progressed significantly over the past decade. Early efforts, such as Pirsiavash \textit{et al.} \cite{pirsiavash2014assessing}, formulated AQA as a direct regression problem, mapping action representations to performance scores. Parisi \textit{et al.} \cite{parisi2016human} instead evaluated action quality by measuring the correctness of action matches.

More recently, research has shifted towards \textit{long-term AQA}, which requires handling extended durations, complex temporal dependencies, and subjective scoring. To address these challenges, Xu \textit{et al.} \cite{xu2022likert} proposed Likert scoring with grade decoupling for more nuanced evaluation, while Zhou \textit{et al.} \cite{zhou2023hierarchical} introduced Hierarchical GCNs to capture structural and temporal hierarchies. CoFInAl \cite{zhou2024cofinal} further improved interpretability by aligning action segments with coarse-to-fine instructions, and PHI \cite{zhou2025phi} tackled domain shift through progressive hierarchical instruction, enhancing robustness across datasets. In parallel, causal inference has been explored to improve robustness and interpretability. FineCausal \cite{han2025finecausal}, for instance, introduced a causal framework for fine-grained AQA. However, such approaches may sacrifice generalizability by overfitting to dataset-specific causal pathways, underscoring the need for methods that balance causal robustness with broad applicability. 

\subsection{Deep Causality Learning}
The integration of causal inference into deep learning, often referred to as Deep Causality Learning, has emerged as a promising direction for improving robustness, interpretability, and generalization. A central idea is the use of \textit{counterfactuals}, which reason about ``what if'' scenarios by altering specific features. For example, Xiao \textit{et al.} \cite{xiao2023masked} introduced masked images as counterfactual samples to enhance fine-tuning robustness, while Rao \textit{et al.} \cite{rao2021counterfactual} proposed counterfactual attention learning to highlight causally relevant cues in fine-grained categorization and person re-identification.

Causal reasoning has also been extended to video understanding. Wei \textit{et al.} \cite{wei2023visual} developed visual causal scene refinement for Video Question Answering (VQA), aiming to disentangle object-event relations, and Liu \textit{et al.} \cite{liu2023cross} modeled cross-modal causal links for video-text reasoning. In perceptual tasks, Shen \textit{et al.} \cite{shen2025image} investigated causal perceptual effects in Image Quality Assessment through abductive counterfactual inference, while Liang \textit{et al.} \cite{liang2024confounded} proposed de-confounded gaze estimation by mitigating spurious correlations. Collectively, these works demonstrate the potential of deep causality learning to explicitly model causal structures and counterfactuals, yielding more robust and interpretable visual models.

\subsection{Flow Matching}
Flow Matching (FM) models represent a cutting-edge advancement in generative modeling. This technique leverages neural Ordinary Differential Equations (ODEs) to implicitly learn a smooth transformation, or “flow,” that maps samples from a simpler base distribution to a more complex target data distribution \cite{tong2023improving}. Building on the principles of normalizing flows \cite{albergo2022building,ma2025uncertainty} and continuous normalizing flows \cite{onken2021ot}, FM models excel at generating high-fidelity samples without the computational overhead of complex approximate inference methods \cite{kobyzev2020normalizing}. A key innovation in recent FM approaches is the development of training algorithms that circumvent the computational difficulties of backpropagating through ODEs, requiring explicit ODE solving only during the inference phase \cite{tong2023improving, lipman2022flow, albergo2022building,liu2022flow,neklyudov2022action}. Consequently, FM presents a highly promising and relatively underexplored direction for generative modeling, offering an efficient and effective means to learn and sample from intricate data distributions. In contrast to diffusion models \cite{ho2020denoising,chang2023design,chang2022unifying,wang2025fg,wang2023fg}, which rely on Stochastic Differential Equations (SDEs) and typically assume a Gaussian base distribution \cite{song2020score}, FM provides enhanced flexibility. It allows for a broader selection of base distributions and employs ODEs for training rather than SDEs, resulting in smoother generative trajectories and often superior performance \cite{song2020denoising}.

\section{Methodology}
\label{sec:formatting}
This section first introduces our proposed framework, followed by a detailed explanation of its core components.

\subsection{Motivation and Framework Overview}
\label{subsec:motivation}

\mypara{Motivation}
Assessing long-term human actions requires models to capture subtle execution details while remaining robust to confounding factors such as environment, background, or recording conditions. However, existing AQA methods face two major limitations. First, they are vulnerable to spurious correlations, where irrelevant contextual cues (e.g., venue or camera angle) become entangled with performance scores, leading to biased predictions. Second, they typically adopt unidirectional temporal modeling, which produces unstable and poorly aligned representations over long sequences, making it difficult to capture the full temporal dynamics of an action. Together, these issues undermine robustness, interpretability, and generalization, limiting the reliability of current AQA approaches. Motivated by these challenges, we introduce \textbf{CaFlow}, a unified framework that integrates counterfactual de-confounding with bidirectional temporal flow, explicitly addressing both the causal and temporal limitations of prior methods.

\mypara{Framework Overview}
CaFlow is designed for Action Quality Assessment (AQA). Given an input action video 
$\mathbf{X}_i \in \mathbb{R}^{T \times W \times H \times 3}$ with $T$ frames of resolution $W \times H$ and 3 color channels, 
the video is first divided into $M$ non-overlapping clips and passed through a pre-trained backbone to obtain an initial feature sequence 
$H_i^0 = \{h_m^0\}_{m=1}^M$, where $h_m^0 \in \mathbb{R}^D$ denotes the feature vector for the $m$-th clip. 
CaFlow then processes these features to predict a scalar action quality score $S \in \mathbb{R}$.
As illustrated in Fig.~\ref{fig:overview}, CaFlow consists of two key components, optimized jointly to achieve robust and generalized AQA:
\begin{enumerate}
    \item \textbf{Causal Counterfactual Regularization (CCR)}: This module introduces a Causal Feature Separator to disentangle causal from confounding clips within the input feature sequence $H^0$. It then applies counterfactual mixing with a triplet-style loss to enforce causally robust representations, ensuring the model focuses on true causal cues for action quality.
    \item \textbf{Bidirectional Time-conditioned Flow (BiT-Flow)}: This time-conditioned bidirectional flow module progressively refines $H^0$ into stable AQA-specific representations $H^1 = \{h_m^1\}_{m=1}^M$ under forward-backward consistency. Here, $h_m^0, h_m^1 \in \mathbb{R}^{D'}$ represent the feature vectors before and after refinement at clip $m$. This refinement ensures temporal consistency and enhances the stability of the learned features.
\end{enumerate}
Finally, the refined representation $H^1$ is aggregated (e.g., via pooling) and regressed by an MLP to predict the final action quality score $S$. 
The overall optimization objective for CaFlow is a high-level combination of these components:
\[
\mathcal{L}_{\text{total}} = \mathcal{L}_{\text{regression}} + \lambda_1 \mathcal{L}_{\text{CCR}} + \lambda_2 \mathcal{L}_{\text{BiT}},
\]
where $\mathcal{L}_{\text{regression}}$ is the primary loss for predicting the action quality score, $\mathcal{L}_{\text{CCR}}$ (see \cref{eq:lccr}) is the loss associated with causal counterfactual regularization, and $\mathcal{L}_{\text{BiT-Flow}}$ (see \cref{eq:bit}) enforces temporal consistency within the flow module. $\lambda_1$ and $\lambda_2$ are hyperparameters balancing these objectives. This comprehensive design compels the model to focus on truly causal cues while ensuring temporally consistent refinement, thereby enhancing robustness and generalization in long-term AQA. Each component will be detailed in the subsequent sections.

\begin{figure}[t] 
    \centering 
    \includegraphics[width=0.35\textwidth]{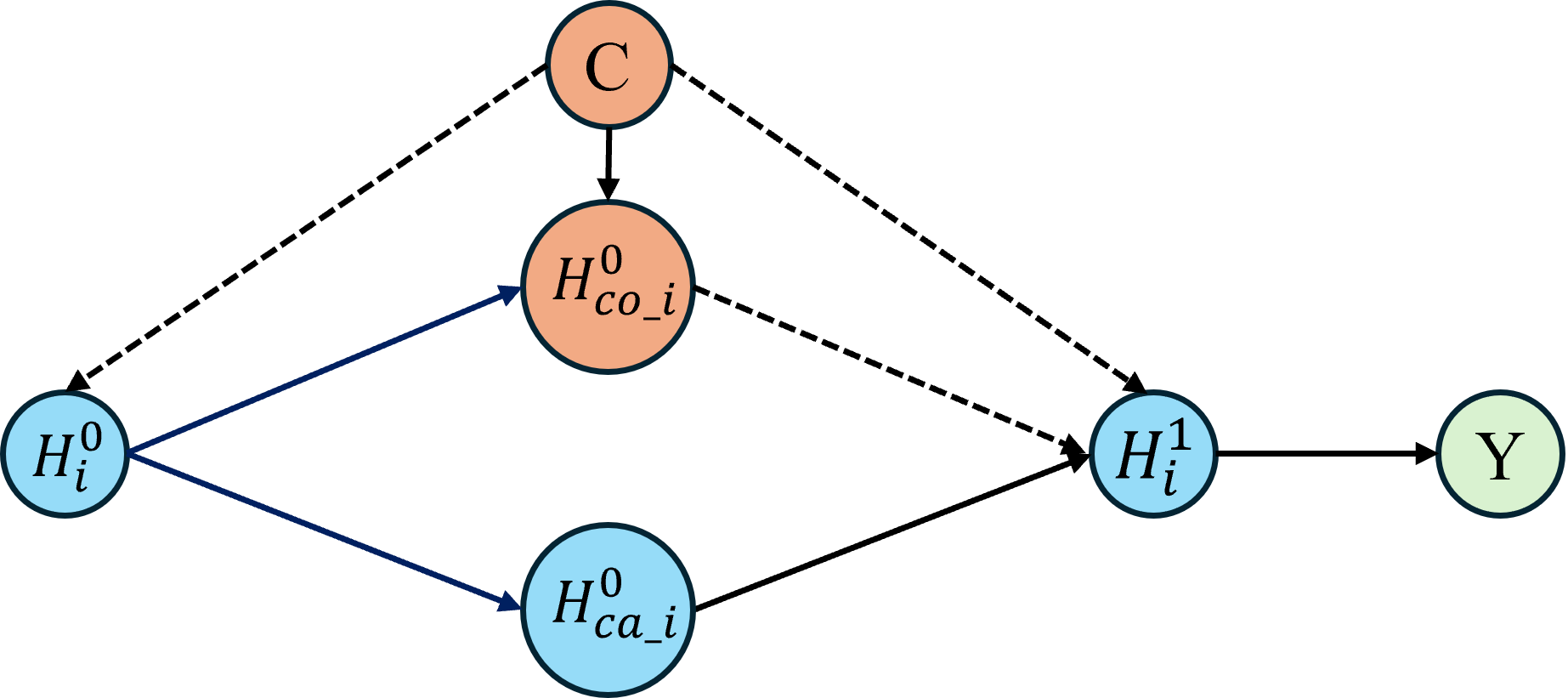} 
    \caption{The causal graph of our AQA framework. Nodes represent variables: 
    \(H_i^0\) for initial video features, \(H_i^1\) for desired features, 
    \(C\) for confounder, \(H_{co\_i}^0\) for confound features, 
    \(H_{ca\_i}^0\) for causal features, and \(Y\) for the final action score. 
    Solid arrows (\(\rightarrow\)) indicate true causal relationships, 
    whereas dashed arrows (\(\dashrightarrow\)) represent spurious causal relationships.} 
    \label{fig:causal_graph}
    \vspace{-1em} 
\end{figure}

\subsection{Causal Counterfactual Regularization}
\mypara{Design Idea}
Existing causal AQA methods \cite{han2025finecausal} reduce spurious correlations by balancing foreground and background features, but their reliance on costly human-annotated masks limits generalizability. Our key insight is to achieve causal separation in a mask-free, self-supervised manner by exploiting the “desired feature” representation from the transformer as an internal guidance signal. This allows our Causal Feature Separator to disentangle causal and confounding features without external supervision. Moreover, instead of implicitly re-weighting features, we explicitly partition them and apply counterfactual testing, forcing the model to reject spurious dependencies. In the following, we present the causal graph, explain the separation mechanism, and introduce the counterfactual intervention strategy.

\mypara{Structural Causal Model}
To model the feature de-confounding process in AQA, we formulate a Structural Causal Model (SCM) as illustrated in Fig.~\ref{fig:causal_graph}. The variables are defined as follows: $H_i^0$ denotes the initial sequence of clip features extracted from the backbone for video $i$; $H_{ca\_i}^0$ and $H_{co\_i}^0$ represent the decomposition of $H_i^0$ into score-causal features and score-confounding features, respectively; $C$ is an unobserved confounder corresponding to environmental factors such as lighting or venue conditions, which may induce spurious correlations; $H_i^1$ is the desired feature representation generated by the transformer module, serving as a proxy for the scoring intent; and $Y$ is the final ground-truth action quality score. The true causal pathway for robust AQA should follow $H_{i}^0 \rightarrow H_{ca\_i}^0 \rightarrow H_i^1 \rightarrow Y$, where the extracted causal features influence the desired representation and ultimately determine the score. However, the confounder \(C\) introduces biased backdoor routes by simultaneously affecting both the raw stream and the desired representation, i.e.,
$H_i^0 \;\leftarrow\; C \;\rightarrow\; H_i^1 \;\rightarrow\; Y$
which create spurious correlations between appearance/context (e.g., venue or lighting) and the final score. In practice, this means that high scores recorded under certain environments can make the model falsely treat those environments as predictive, even when execution quality is unchanged.

To focus on the true causal effect $H_i^0 \!\to\! H_{ca_i}^0 \!\to\! H_i^1 \!\to\! Y$ and eliminate the non-causal impact of the backdoor paths opened by $C$, we adopt a front-door intervention that treats $H_{ca_i}^0$ as the mediator. Since $H_{ca_i}^0$ intercepts all directed paths from $H_i^0$ to $Y$ and is not directly influenced by $C$, the causal effect of $H^0$ on $Y$ can be estimated via:
\begin{equation}
\begin{aligned}
P(Y \mid do(H^0=h)) 
= & \sum_{h_{\mathrm{ca}}} P(h_{\mathrm{ca}} \mid h) \times \sum_{h^1} P(h^1 \mid h_{\mathrm{ca}}) \times \\
  & \sum_{h'} P(Y \mid h^1, h')\,P(h'),
\end{aligned}
\label{eq:fd} 
\end{equation}
where $h'$ indexes raw features from the observational distribution. 
This formulation blocks the spurious influences of $C \!\to\! H^0$ 
and $C \!\to\! H^1$, ensuring that the model learns only from the 
true causal pathway.

\mypara{Causal Feature Separator (CFS)}
Inspired by recent advances in causal discovery through attention mechanisms \cite{wei2023visual}, we design a Causal Feature Separator (CFS) to dynamically divide the initial feature sequence $H_i^0$ into a score-causal subset $H_{\mathrm{causal}}$ and a score-confounding subset $H_{\mathrm{confound}}$ in a self-supervised fashion. The separation is guided by the “desired feature” $H_i^1$ obtained from the temporal transformer, which acts as a high-level proxy for scoring intent. To realize this, we compute attention scores $\mathbf{a}_i \in \mathbb{R}^M$ through a cross-attention operation, where $H_i^1$ serves as the \textit{Query} and the individual clip features $\{h_{i,t}^0\}_{t=1}^M$ from $H_i^0$ act as the \textit{Keys} and \textit{Values}. These attention scores measure the contribution of each clip feature to the desired representation, enabling the model to separate causal cues from confounding context.

To separate causal and confounding features, we apply the Gumbel-Softmax function to the continuous attention scores. This technique injects stochasticity through Gumbel noise and employs a temperature-controlled softmax to produce near-binary probability weights. Formally, we obtain a soft mask $\mathbf{m}_i \in [0,1]^M$, which is:
\begin{equation}
    \mathbf{m}_i = \text{Gumbel-Softmax}(\mathbf{a}_i, \tau),
\end{equation}
where $\tau$ is the temperature parameter annealed during training. Each element of $\mathbf{m}_i$ reflects the probability of the corresponding clip feature being causal. Using this mask, the initial feature sequence $H_i^0$ is partitioned into two complementary components:
\begin{align}
    H_{\mathrm{causal}} &= \mathbf{m}_i \odot H_i^0,
    H_{\mathrm{confound}} = (1 - \mathbf{m}_i) \odot H_i^0,
\end{align}
where $\odot$ denotes element-wise multiplication. This procedure enables the model to softly separate causal and confounding features in a differentiable manner, without requiring external supervision.

\mypara{Counterfactual Regularization via Causal Distillation.}
Drawing from the principles of counterfactual sample generation, we create counterfactuals in the feature space to explicitly break spurious correlations and regularize the model \cite{xiao2023masked}. For a video $i$ and another randomly selected video $h$ from the same training batch, we first generate two types of counterfactual feature sequences by swapping their respective causal and confounding components:
\begin{enumerate}
    \item \textbf{Confounder-Swapped Counterfactual ($H_{\mathrm{cf\_conf}}$):} We combine the causal features from video $i$ with the confounding features from video $h$: $H_{\mathrm{cf\_conf}} = H_{\mathrm{causal}\_i} \cup H_{\mathrm{confound}\_h}$. This sample retains the original causal information but introduces a new, potentially conflicting confounding context.
    \item \textbf{Causal-Swapped Counterfactual ($H_{\mathrm{cf\_causal}}$):} We combine confounding features from video $i$ with causal features from video $h$: $H_{\mathrm{cf\_causal}} = H_{\mathrm{confound}\_i} \cup H_{\mathrm{causal}\_h}$. This sample retains original confounding context but replaces core causal information.
\end{enumerate}
With these generated counterfactuals, we introduce a causal distillation objective, $\mathcal{L}_{\text{CCR}}$, designed to make the model invariant to confounding information while remaining sensitive to causal information. Let $Z(H)$ denote the refined feature output by the model's primary feature encoder. This encoder is identical to the one used in the temporal transformer module, ensuring consistency in the feature space across both causal separation and temporal modeling. We formulate $\mathcal{L}_{\text{CCR}}$ using a triplet-style objective:
\begin{equation} \label{eq:lccr}
\begin{aligned}
    \mathcal{L}_{\text{CCR}} = \max(0, D_{pos} - D_{neg} + m), \\
    \text{where } D_{pos} = D(Z(H^0_i), Z(H_{\mathrm{cf\_conf}})), \\
    \text{and } D_{neg} = D(Z(H^0_i), Z(H_{\mathrm{cf\_causal}})). 
\end{aligned}
\end{equation}
where $D$ is a distance metric like Mean Squared Error (MSE), and $m$ is a margin hyperparameter. This loss encourages the distance between the representations of the original and the confounder-swapped features to be small, as their causal core is unchanged. Simultaneously, it pushes the distance between the original and the causal-swapped features to be large, as their causal core has been replaced. This directly produces representations that rely on the identified causal features and ignore the confounding ones.

\subsection{Time-Conditioned Bidirectional Flow}
\label{sec:bitflow}

\mypara{Design Idea}
Standard flow-matching techniques in AQA often suffer from unstable representation refinement and temporal misalignment, as they model the process in a purely unidirectional manner \cite{zhou2025phi}. To address this limitation, we draw inspiration from the Schrödinger Bridge problem \cite{tong2023simulation}, which provides a bidirectional formulation that simultaneously satisfies boundary conditions at both the start and end points. Building on this principle, BiT-Flow explicitly models both forward and backward temporal dynamics and enforces them to act as approximate inverses through a cycle-consistency constraint. Unlike unidirectional flow models that refine representations by repeatedly pushing them forward, BiT-Flow produces smoother and more coherent representation trajectories by coupling time-aware refinement with bidirectional symmetry. This design stabilizes training, prevents divergence, and yields temporally aligned features that are better suited for capturing fine-grained execution details in AQA.

\mypara{Implementation}
The BiT-Flow module is implemented with three components. 
First, for time conditioning, a normalized time step \(t \in [0,1]\) is embedded using an MLP,
\begin{equation}
\mathbf{e}(t) = \mathrm{MLP}(t),
\end{equation}
and injected into both the encoder inputs \(\mathbf{X}\) and the decoder queries \(\mathbf{Q}\) to condition the refinement process on its temporal stage, which can be represented as:
\begin{align}
\tilde{\mathbf{X}} &= \mathbf{X} + \mathbf{e}(t),\tilde{\mathbf{Q}} = \mathbf{Q} + \mathrm{mean}_{n}\!\big(\mathbf{e}(t)\big).
\end{align}
Second, for bidirectional flow, two direction-specific predictors generate forward and backward flows, which are blended by a schedule \(\alpha(t)\):
\begin{equation}
\Delta\mathbf{H}^{\mathrm{fwd}}_t = F_{\mathrm{fwd}}(\tilde{\mathbf{X}}, \tilde{\mathbf{Q}}, t),
\Delta\mathbf{H}^{\mathrm{bwd}}_t = F_{\mathrm{bwd}}(\tilde{\mathbf{X}}, \tilde{\mathbf{Q}}, 1{-}t),
\end{equation}
\begin{equation}
\Delta\mathbf{H}_t = \alpha(t)\,\Delta\mathbf{H}^{\mathrm{fwd}}_t + (1{-}\alpha(t))\,\Delta\mathbf{H}^{\mathrm{bwd}}_t.
\end{equation}
Finally, to ensure forward-backward consistency, we introduce losses that regularize the flow updates. With a prototype anchor \(\mathbf{P}\), the flow loss aligns the blended update with the target change:
\begin{equation}
\mathcal{L}_{\mathrm{flow}} = \|\Delta\mathbf{H}_t - (\mathbf{H}_{t+1}-\mathbf{P})\|_2^2,
\end{equation}
while the consistency loss enforces a cycle:
\begin{equation}
\begin{aligned}
\mathcal{L}_{\mathrm{cons}} =& \|\mathbf{P} + \Delta\mathbf{H}_t - \mathbf{H}_{t+1}\|_2^2 \\
&+ \|\mathbf{H}_{t+1} + \Delta\mathbf{H}^{\mathrm{bwd}}_{t} - \mathbf{P}\|_2^2.
\end{aligned}
\end{equation}
The final BiT-Flow objective combines both terms:
\begin{equation} \label{eq:bit}
\mathcal{L}_{\mathrm{BiT}} = \mathcal{L}_{\mathrm{flow}} + \lambda_{\mathrm{cons}}\,\mathcal{L}_{\mathrm{cons}}.
\end{equation}

The forward-backward consistency loss \(\mathcal{L}_{\mathrm{cons}}\) is the critical innovation of BiT-Flow, acting as a powerful regularizer that forces the two flows to be approximate inverses. This stabilizes training, prevents erratic updates, and produces smooth, temporally coherent representation trajectories. For AQA, where subtle variations in execution determine quality, such enforced smoothness is essential. By coupling time-aware refinement with bidirectional symmetry, BiT-Flow delivers more stable supervision, improved temporal alignment, and more accurate quality assessment.

\begin{table*}[t]
\centering
\caption{Results of \SRCC{} (\(\uparrow\)) and \RLtwo{} (\(\downarrow\)) on the RG dataset. Best results are in \textbf{bold}, second best are \second{underlined}. “$^{\ast}$” indicates our reimplementation based on the official code. Average \SRCC{} is computed using Fisher-\emph{z}. “--”: not reported, “+”: extra features/modalities.}
\label{tab:rg_results}
\setlength{\tabcolsep}{3.6pt}
\renewcommand{\arraystretch}{1.18}
\scalebox{0.75}{%
\begin{tabular}{l l l l
                *{2}{r} *{2}{r} *{2}{r} *{2}{r} *{2}{r}}
\toprule
\multirow{2.5}{*}{Method} & \multirow{2.5}{*}{Publisher} & \multirow{2.5}{*}{Backbone} & \multirow{2.5}{*}{Modality}
& \multicolumn{2}{c}{Ball}
& \multicolumn{2}{c}{Clubs}
& \multicolumn{2}{c}{Hoop}
& \multicolumn{2}{c}{Ribbon}
& \multicolumn{2}{c}{Average} \\
\cmidrule(lr){5-6}\cmidrule(lr){7-8}\cmidrule(lr){9-10}\cmidrule(lr){11-12}\cmidrule(lr){13-14}
& & & & \SRCC & \RLtwo & \SRCC & \RLtwo & \SRCC & \RLtwo & \SRCC & \RLtwo & \SRCC & \RLtwo \\
\midrule
MS-LSTM \cite{xu2019learning}      & TCSVT'19  & VST        & RGB   & 0.621 & --    & 0.661 & --    & 0.670 & --    & 0.695 & --    & 0.663 & --    \\
ACTION-NET \cite{zeng2020hybrid}   & ACM MM'20 & VST$^{+}$  & RGB   & 0.684 & --    & 0.737 & --    & 0.733 & --    & 0.754 & --    & 0.728 & --    \\
GDLT \cite{xu2022likert}           & CVPR'22   & VST        & RGB   & 0.746 & 2.833 & 0.802 & 2.179 & 0.765 & \best{2.012} & 0.741 & \second{2.579} & 0.765 & \second{2.401} \\
HGCN$^{\ast}$ \cite{zhou2023hierarchical} & TCSVT'23  & VST        & RGB   & 0.711 & 3.030 & 0.789 & 3.444 & 0.728 & 5.312 & 0.703 & 5.576 & 0.735 & 4.341 \\
PAMFN \cite{zeng2024multimodal}    & TIP'24    & VST        & RGB   & 0.636 & --    & 0.720 & --    & 0.769 & --    & 0.708 & --    & 0.711 & --    \\
VATP-Net \cite{gedamu2024visual}   & TCSVT'24  & VST        & RGB+  & 0.800 & --    & \second{0.810} & --    & 0.780 & --    & 0.769 & --    & 0.800 & --    \\
CoFInAl  \cite{zhou2024cofinal}    & IJCAI'24  & VST        & RGB   & 0.809 & \best{1.356} & 0.806 & 2.453 & 0.804 & 9.918 & \second{0.810} & \best{2.383} & \second{0.807} & 4.028 \\
PHI \cite{zhou2025phi}             & TIP'25    & VST        & RGB   & \second{0.818} & 2.187 & 0.803 & \second{2.149} & \second{0.812} & \second{2.119} & 0.805 & 2.744 & 0.810 & \best{2.300} \\
CaFlow (Ours)                      & --        & VST        & RGB   & \best{0.863} & \second{2.101} & \best{0.822} & \best{2.048} & \best{0.843} & 2.985 & \best{0.833} & 2.663 & \best{0.841} & 2.449 \\
\bottomrule
\end{tabular}}
\end{table*}

\begin{table*}[t]
\centering
\caption{Results of \SRCC{} (\(\uparrow\)) and \RLtwo{} (\(\downarrow\)) on the FIS-V dataset. 
Best results are in \textbf{bold} and second best are \second{underlined}. 
“$^{\ast}$” indicates our reimplementation based on the official code. 
Average \SRCC{} uses Fisher-\emph{z}. “--”: not reported, “+”: extra features/modalities.}
\label{tab:fisv_results}
\setlength{\tabcolsep}{3.6pt}
\renewcommand{\arraystretch}{1.18}
\scalebox{0.75}{%
\begin{tabular}{l l l l *{2}{r} *{2}{r} *{2}{r}}
\toprule
\multirow{2.5}{*}{Method} & \multirow{2.5}{*}{Publisher} & \multirow{2.5}{*}{Backbone} & \multirow{2.5}{*}{Modality}
& \multicolumn{2}{c}{TES}
& \multicolumn{2}{c}{PCS}
& \multicolumn{2}{c}{Average} \\
\cmidrule(lr){5-6}\cmidrule(lr){7-8}\cmidrule(lr){9-10}
& & & & \SRCC & \RLtwo & \SRCC & \RLtwo & \SRCC & \RLtwo \\
\midrule
MS-LSTM \cite{xu2019learning}      & TCSVT'19  & VST       & RGB   & 0.660 & --    & 0.809 & --    & 0.744 & --    \\
ACTION-NET \cite{zeng2020hybrid}   & ACM MM'20 & VST$^{+}$ & RGB   & 0.694 & --    & 0.809 & --    & 0.757 & --    \\
GDLT \cite{xu2022likert}           & CVPR'22   & VST       & RGB   & 0.685 & 3.717 & 0.820 & 2.072 & 0.761 & 2.895 \\
HGCN$^{\ast}$ \cite{zhou2023hierarchical} & TCSVT'23  & VST       & RGB   & 0.246 & 12.628& 0.221 & 20.531& 0.234 & 16.580 \\
MLP-Mixer \cite{xia2023skating}    & AAAI'23   & VST       & RGB   & 0.680 & --    & 0.820 & --    & 0.750 & --    \\
SGN \cite{du2024learning}          & TMM'24    & VST       & RGB   & 0.700 & --    & 0.830 & --    & 0.765 & --    \\
PAMFN \cite{zeng2024multimodal}    & TIP'24    & VST       & RGB   & 0.665 & --    & 0.823 & --    & 0.755 & --    \\
VATP-Net \cite{gedamu2024visual}   & TCSVT'24  & VST       & RGB+  & 0.702 & --    & 0.863 & --    & 0.796 & --    \\
CoFInAl  \cite{zhou2024cofinal}    & IJCAI'24  & VST       & RGB   & 0.716 & 2.875 & 0.843 & 1.752 & 0.788 & 2.314 \\
PHI \cite{zhou2025phi}             & TIP'25    & VST       & RGB   & \second{0.726} & \second{2.543} & \best{0.867} & \second{1.656} & \second{0.804} & \second{2.178} \\
CaFlow (Ours)                      & --        & VST       & RGB   & \best{0.729} & \best{2.480} & \second{0.865} & \best{1.619} & \best{0.805} & \best{2.050} \\
\bottomrule
\end{tabular}}
\end{table*}

\begin{table}[t]
\centering
\caption{Results of SRCC (\(\uparrow\)) and \RLtwo{} (\(\downarrow\)) on the LOGO dataset.
Best results are in \textbf{bold}, second best are \second{underlined}.}
\label{tab:logo_results}
\setlength{\tabcolsep}{6pt}
\renewcommand{\arraystretch}{1.15}
\resizebox{\linewidth}{!}{%
\begin{tabular}{l l l l c c}
\toprule
Method & Publisher & Backbone & Modality & SRCC & \RLtwo \\
\midrule
USDL \cite{tang2020uncertainty}        & CVPR'20  & VST & RGB & 0.530 & 4.997 \\
CoRe \cite{yu2021group}                & ICCV'21  & VST & RGB & 0.503 & 5.596 \\
TSA \cite{xu2022finediving}            & CVPR'22  & VST & RGB & 0.570 & 4.536 \\
CoRe-GOAT \cite{zhang2023logo}         & CVPR'23  & VST & RGB & 0.574 & 4.437 \\
HGCN \cite{zhou2023hierarchical}       & TCSVT'23 & VST & RGB & 0.475 & 4.640 \\
CoFInAl  \cite{zhou2024cofinal}        & IJCAI'24 & VST & RGB & 0.698 & 4.019 \\
PHI \cite{zhou2025phi}                 & TIP'25   & VST & RGB & \second{0.835} & \second{2.752} \\
CaFlow (Ours)                          & --       & VST & RGB & \best{0.856} & \best{1.425} \\
\bottomrule
\end{tabular}}
\vspace{-0.5cm}
\end{table}

\section{Experiments}
This section first introduces the experimental setup and then presents and analyzes the results.

\subsection{Experimental Setting}
\mypara{Datasets}
Our evaluation covers three widely used benchmarks for long-term AQA. The first is the \textbf{Rhythmic Gymnastics (RG)} dataset \cite{zeng2020hybrid}, which contains 1,000 clips of athletes performing routines with four apparatus types: Ball, Clubs, Hoop, and Ribbon. Each sequence lasts roughly 1.6 minutes at 25 fps, and the official protocol allocates 200 training and 50 testing samples per discipline. The second benchmark, the \textbf{Figure Skating Video (Fis-V)} dataset \cite{pirsiavash2014assessing,parmar2017learning}, consists of 500 ladies’ singles short programs averaging 2.9 minutes in length. Following the standard split, 400 samples are reserved for training and 100 for evaluation. Each program is annotated with two types of scores, namely the Total Element Score (TES) and the Program Component Score (PCS). In line with previous work \cite{xu2019learning}, we train separate predictors for the two score categories. The third dataset, \textbf{LOng-form GrOup (LOGO)} \cite{zhang2023logo}, comprises 150 training and 50 test videos of synchronized swimming. With an average duration of 3.5 minutes per sequence, LOGO currently provides the longest video samples among AQA datasets and represents a particularly demanding benchmark for long-term assessment.


\mypara{Evaluation Metrics}
To assess the effectiveness of the proposed method, we adopt two kinds of metrics. 

In line with prior studies on long-term AQA \cite{xu2022likert,zeng2020hybrid}, we employ Spearman’s Rank Correlation Coefficient (SRCC), denoted as $\rho$, which evaluates the monotonic agreement between predictions and ground-truth scores. SRCC is defined as the Pearson correlation between the ranking of predictions $r(\hat{s}_i)$ and the ranking of true labels $r(s_i)$:
\begin{equation} \label{eq_srcc}
    \rho=\frac{\sum_{i=1}^N \big(r(s_i)-\bar{r}\big)\big(r(\hat{s}_i)-\bar{r}\big)}{\sqrt{\sum_{i=1}^N\big(r(s_i)-\bar{r}\big)^2}\;\sqrt{\sum_{i=1}^N\big(r(\hat{s}_i)-\bar{r}\big)^2}},
\end{equation}
where $\bar{r}$ is the mean rank. A larger $\rho$ implies stronger consistency between predicted and true ranking orders. Following \cite{pan2019action}, we compute the average SRCC across different types in RG and across TES/PCS scores in Fis-V using Fisher’s $z$-transformation to combine individual results.

Beyond correlation-based evaluation, we also report a stricter error measure, the relative $\ell_2$ distance (R-$\ell_2$) \cite{yu2021group,zhou2023hierarchical,zhou2024cofinal}. This metric captures the normalized discrepancy between predicted and ground-truth scores, which makes the comparison invariant to the absolute score range. Given the maximum and minimum reference scores $s_{\mathrm{max}}$ and $s_{\mathrm{min}}$, the R-$\ell_2$ distance is calculated as
\begin{equation}
    \text{R-}\ell_2 = \frac{1}{N}\sum_{n=1}^N \left(\frac{|s_n - \hat{s}_n|}{s_{\mathrm{max}} - s_{\mathrm{min}}}\right)^2 \times 100,
\end{equation}
where $s_n$ and $\hat{s}_n$ denote the ground-truth and predicted scores of the $n$-th sample, respectively. For datasets with multiple action or score categories, the final performance is aggregated using Fisher’s $z$-value.

\mypara{Implementation Details}
All experiments are implemented in PyTorch and conducted on an RTX 4080 GPU. Video frames are sampled at 25 fps and uniformly divided into non-overlapping clips of 32 frames. A Vision Swin Transformer (VST) pre-trained on Kinetics-400 is adopted as the feature backbone, producing 1024-dimensional clip embeddings. During training, the start segment is randomly selected, with the number of clips set to $M{=}68$ for RG, $M{=}124$ for Fis-V, and $M{=}48$ for LOGO, respectively. We use the Adam optimizer with an initial learning rate of $1\times10^{-2}$, weight decay of $1\times10^{-4}$, and a batch size of 32. The learning rate is decayed by a factor of 0.1 after 50\% and 75\% of total epochs. To further optimize the networks, we apply a dropout of 0.3. For hyperparameters, the loss balancing factors $\lambda_1$ and $\lambda_2$ are set to 0.02 and 0.5, respectively, while the Gumbel-Softmax temperature is annealed from 1.0 during training. At inference time, clip features are aggregated through average pooling, and the final quality score is obtained from the regression head without any test-time augmentation. As shown in Table 1 of the supplementary material, CaFlow introduces additional parameters only in the offline stage while keeping the online stage lightweight and computationally efficient, yet still surpasses all prior long-term AQA methods with a significant improvement in average SRCC.

\subsection{Comparisons with State-of-the-Arts}
We compare CaFlow against a wide range of state-of-the-art AQA methods on the RG, FIS-V, and LOGO datasets, with results summarized in Tables~\ref{tab:rg_results}--\ref{tab:logo_results}. Across all three benchmarks, CaFlow consistently achieves the best or second-best results, demonstrating its effectiveness and robustness.  

On the RG dataset (Table~\ref{tab:rg_results}), CaFlow achieves the highest overall \SRCC{} of 0.838, outperforming the previous best PHI \cite{zhou2025phi} (0.810) by +3.5\%. For error, CaFlow obtains an average \RLtwo{} of 2.455, which is slightly higher than PHI (2.300), but still represents a 39.9\% reduction compared to earlier methods such as HGCN (4.341). Looking into the four apparatuses, CaFlow demonstrates consistent advantages. For \emph{Ball}, our method achieves an \SRCC{} of 0.863, which improves upon PHI (0.818) by +5.5\%, and reduces the \RLtwo{} from 3.030 (HGCN) to 2.101, a 30.6\% error reduction. For \emph{Clubs}, CaFlow reaches 0.822 in \SRCC{}, surpassing the previous best (0.810 by VATP-Net) by +1.5\%, while also delivering the lowest \RLtwo{} (2.048), reducing error by 40.6\% compared to GDLT (2.179). For \emph{Hoop}, CaFlow attains the best \SRCC{} of 0.843, a +3.8\% gain over PHI (0.812), though its \RLtwo{} of 2.985 lags behind GDLT (2.012) and PHI (2.119). This suggests our method is particularly strong in correlation capture, even if some error variance remains. For \emph{Ribbon}, CaFlow achieves 0.833 in \SRCC{}, outperforming PHI (0.805) by +3.5\%, while its \RLtwo{} of 2.663 is close to PHI (2.744).

On the FIS-V dataset (Table~\ref{tab:fisv_results}), CaFlow achieves the best overall \SRCC{} of 0.805, marginally surpassing PHI (0.804). While the correlation gain appears small (+0.1\%), CaFlow reduces the average \RLtwo{} to 2.050, a 5.9\% improvement over PHI (2.178) and a 29.1\% reduction relative to GDLT (2.895). In terms of event-level performance, CaFlow achieves the best TES score (0.729/2.480), improving \RLtwo{} by 33.3\% over PHI (2.543), while remaining highly competitive on PCS (0.865 vs. 0.867). This demonstrates that our bidirectional flow enhances stability even when correlation scores converge at the top end.  

On the LOGO dataset (Table~\ref{tab:logo_results}), CaFlow sets a new state of the art, achieving an \SRCC{} of 0.856 and \RLtwo{} of 1.425. Compared to PHI (0.835/2.752), this corresponds to a +2.5\% improvement in correlation and a substantial 48.2\% reduction in error. These large margins highlight the strength of explicitly combining counterfactual de-confounding with bidirectional temporal refinement, especially in handling long and complex sequences.  

Overall, CaFlow delivers consistent and significant improvements across datasets. Even when \SRCC{} margins over the strongest baselines are modest, CaFlow consistently achieves substantial reductions in \RLtwo{}, underscoring its robustness in mitigating spurious correlations and stabilizing temporal dynamics for fine-grained AQA.

\begin{table*}[t]
\centering
\caption{Ablation on RG: \SRCC{} (\(\uparrow\)) and \RLtwo{} (\(\downarrow\)) per apparatus.
Average \SRCC{} uses Fisher-\emph{z}. Best results in \textbf{bold}, second best \second{underlined}.}
\label{tab:rg_ablation_full}

\setlength{\tabcolsep}{3.6pt}
\scalebox{0.8}{%
\begin{tabular}{l c c *{2}{c} *{2}{c} *{2}{c} *{2}{c} *{2}{c}}
\toprule
\multirow{2.5}{*}{Variant} &
\multirow{2.5}{*}{CCR} &
\multirow{2.5}{*}{BiT} &
\multicolumn{2}{c}{Ball} &
\multicolumn{2}{c}{Clubs} &
\multicolumn{2}{c}{Hoop} &
\multicolumn{2}{c}{Ribbon} &
\multicolumn{2}{c}{Average} \\

\cmidrule(lr){4-5}
\cmidrule(lr){6-7}
\cmidrule(lr){8-9}
\cmidrule(lr){10-11}
\cmidrule(lr){12-13}
& & &
\SRCC & \RLtwo &
\SRCC & \RLtwo &
\SRCC & \RLtwo &
\SRCC & \RLtwo &
\SRCC & \RLtwo \\
\midrule

Backbone + Regressor (Baseline) & \xmark & \xmark
& 0.818 & \second{2.187}
& 0.803 & 2.149
& 0.812 & \best{2.119}
& 0.805 & 2.744
& 0.810 & \best{2.300} \\

+ CCR & \cmark & \xmark
& \second{0.843} & 2.855
& \second{0.821} & \best{1.950}
& \second{0.836} & 3.791
& \best{0.843} & \best{2.109}
& \second{0.836} & 2.676 \\

+ BiT-Flow & \xmark & \cmark
& 0.833 & 3.052
& 0.806 & 2.271
& 0.829 & 5.619
& 0.827 & 5.245
& 0.824 & 4.047 \\

CCR + BiT-Flow (CaFlow, Ours) & \cmark & \cmark
& \best{0.863} & \best{2.101}
& \best{0.822} & \second{2.048}
& \best{0.843} & \second{2.985}
& \second{0.833} & \second{2.663}
& \best{0.841} & \second{2.449} \\

\bottomrule
\end{tabular}}
\end{table*}

\begin{table}[t]
\centering
\caption{Ablation on LOGO: \SRCC{} (\(\uparrow\)) and \RLtwo{} (\(\downarrow\)). Best results in \textbf{bold}, second best \second{underlined}.}
\label{tab:logo_ablation}
\setlength{\tabcolsep}{6pt}
\renewcommand{\arraystretch}{1.15}
\scalebox{0.8}{%
\begin{tabular}{l c c c c}
\toprule
Variant & CCR & BiT & SRCC & \RLtwo \\
\midrule
Backbone + Regressor (Baseline) & \xmark & \xmark & 0.835 & 2.752 \\
+ CCR                           & \cmark & \xmark & \second{0.849} & \second{1.916} \\
+ BiT\textendash Flow            & \xmark & \cmark & 0.845 & 2.179 \\
CCR + BiT\textendash Flow (CaFlow, Ours) & \cmark & \cmark & \best{0.856} & \best{1.425} \\
\bottomrule
\end{tabular}}
\end{table}

\subsection{Ablation Study}
We evaluate the individual contributions of CCR and BiT-Flow on the RG and LOGO datasets. On RG (Table~\ref{tab:rg_ablation_full}), both modules outperform the baseline. CCR consistently boosts correlations, especially on \emph{Ball} (+3.1\%) and \emph{Ribbon} (+4.7\%), validating its ability to disentangle causal cues. BiT-Flow alone stabilizes features but yields mixed results, with limited \SRCC{} gains and higher \RLtwo{}. When combined, CaFlow achieves the best average \SRCC{} (0.841) with competitive error (2.449), showing the two modules are complementary. On LOGO (Table~\ref{tab:logo_ablation}), CCR reduces error by 30.4\% and raises \SRCC{} by +1.7\%. BiT-Flow also improves over baseline, while their combination sets a new state-of-the-art with \SRCC{} of 0.856 and \RLtwo{} of 1.425. We omit ablations on FIS-V since improvements over the strong PHI baseline are small, with gains mainly emerging when both modules are integrated, consistent with RG and LOGO.

\subsection{Qualitative Analysis}
\begin{figure}[t]
  \centering
  \includegraphics[width=\linewidth]{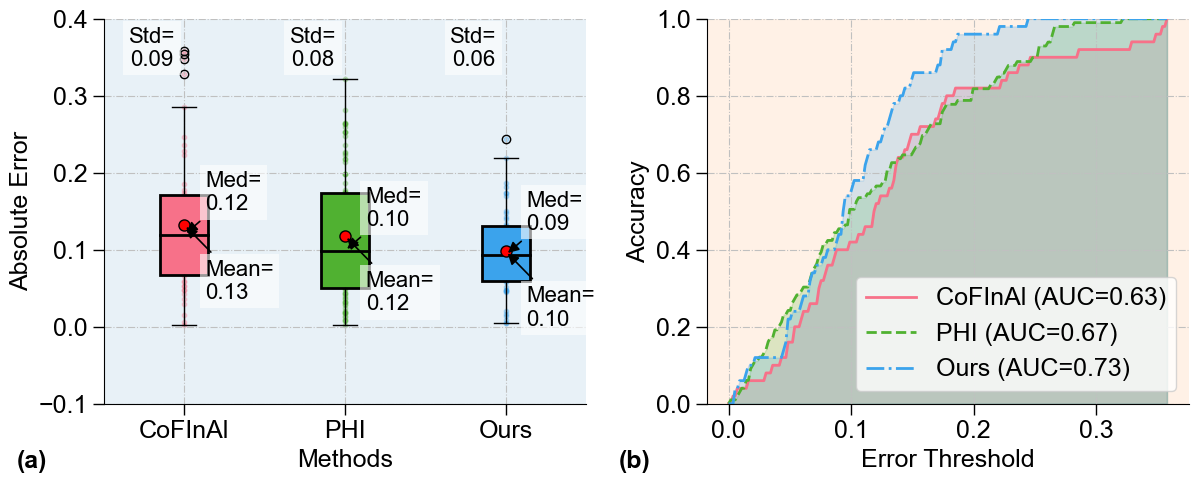}
  \caption{Error analysis on RG. 
  (a) Boxplots of absolute errors with annotated statistics (mean/median/std). 
  (b) Cumulative error-accuracy curves with area under the curve (AUC).}
  \label{fig:auc_box}
  \vspace{-1em}  
\end{figure}
Fig.~\ref{fig:auc_box}(a) shows that our method yields the lowest dispersion and bias among the three systems: the mean absolute error drops to 2.44 versus 2.95 for PHI and 3.31 for CoFInAI; the median decreases to 2.34 (PHI: 2.47, CoFInAI: 2.98); and the standard deviation narrows to 1.38 (PHI: 2.01, CoFInAI: 2.23). 
The cumulative accuracy curves in Figure~\ref{fig:auc_box}(b) further confirm this trend: CaFlow attains the largest AUC of 0.73, compared to 0.67 for PHI and 0.63 for CoFInAI, with a clear advantage at small error thresholds—precisely the regime needed for reliable judging.

\begin{figure}[t]
  \centering
  \includegraphics[width=0.9\linewidth]{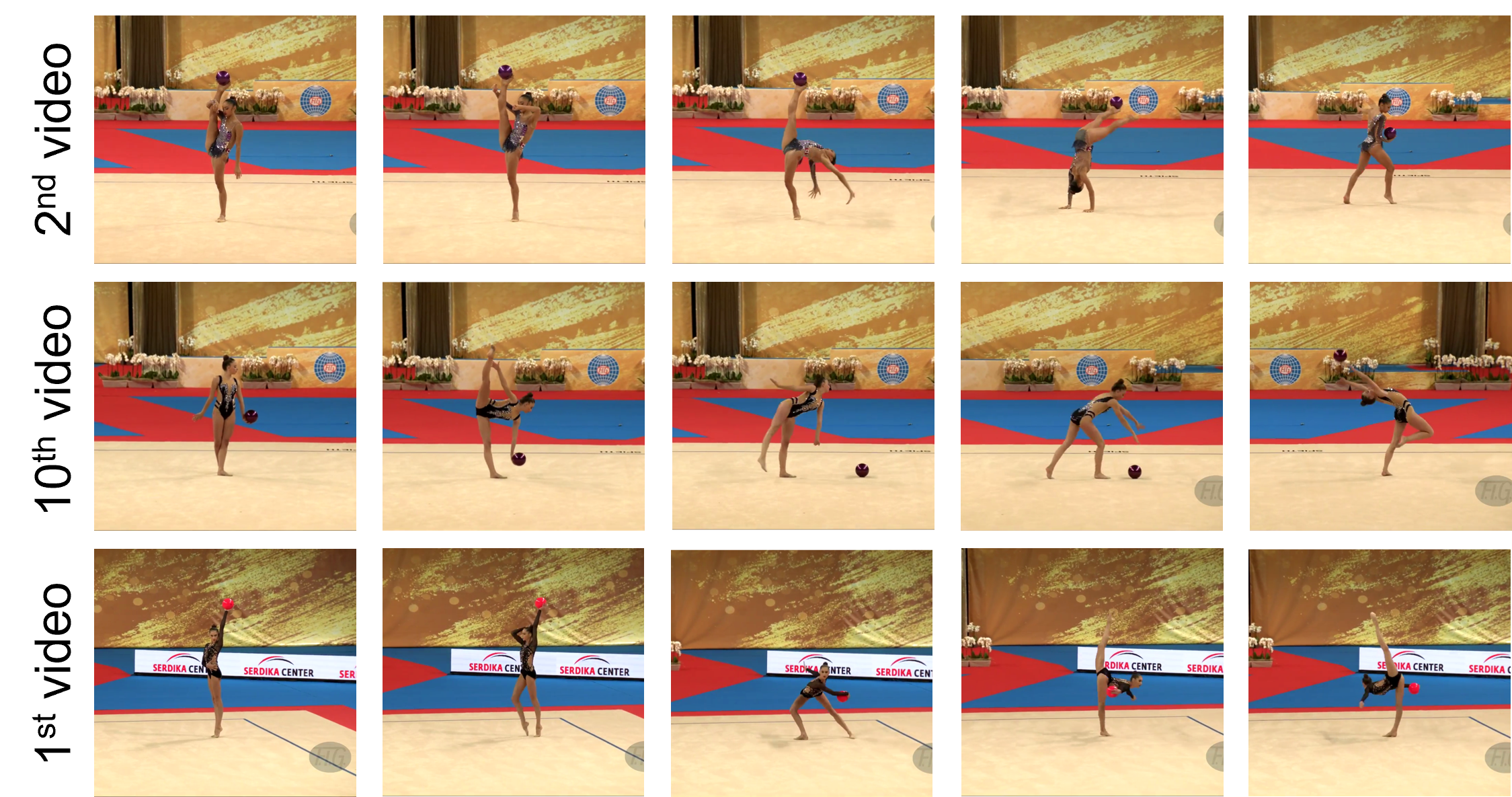}
  \caption{Three representative routines with key frames per case.}
  \label{fig:qual_cases}
  \vspace{-1em}  
\end{figure}

Fig.~\ref{fig:qual_cases} presents three representative videos highlighting the comparative strengths and weaknesses of the evaluated methods. 
In the first case (Video~2, GT $16.70$), CoFInAI underestimates by $14.90$, PHI also underestimates at $15.79$, while CaFlow predicts $17.57$, closest to the ground truth, showing its ability to capture subtle execution details. 
In the second case (Video~10, GT $13.65$), CoFInAI gives $11.80$ and PHI $12.53$, both lower than the truth, whereas CaFlow outputs $13.53$, nearly identical to the ground truth, demonstrating robustness to confounding context. 
In the third case (Video~12, GT $16.70$), all methods fail: CoFInAI predicts $15.10$, PHI $18.07$, and CaFlow $19.53$, suggesting that highly complex or ambiguous executions remain challenging. 
Overall, these examples show that CaFlow generally delivers more reliable predictions than prior methods, though extreme cases still leave room for improvement.

\section{Conclusion and Discussion}
We presented \textbf{CaFlow}, a unified framework for Action Quality Assessment that integrates counterfactual de-confounding and bidirectional temporal refinement. Specifically, the \textit{Causal Counterfactual Regularization (CCR)} module disentangles causal from confounding features in a self-supervised manner and enforces robustness through counterfactual interventions, while the \textit{BiT-Flow} module models forward and backward temporal dynamics with cycle consistency, yielding smooth and coherent representation trajectories. Together, these components enable CaFlow to achieve state-of-the-art performance across multiple long-term AQA benchmarks.  

Despite these advances, limitations remain. CCR removes spurious correlations without external supervision, yet separation still depends on internal representations, which may bias under extreme distribution shifts. BiT-Flow stabilizes refinement but adds computational overhead compared to lightweight unidirectional methods. Future work should explore more efficient flow architectures, lightweight regularization, and counterfactual reasoning in low-annotation or semi-supervised settings. Overall, CaFlow offers a robust, interpretable foundation for long-term AQA, showing the benefits of combining causal inference with bidirectional temporal modeling while paving the way for more generalizable, efficient frameworks.



{
    \small
    \bibliographystyle{ieeenat_fullname}
    \bibliography{main}
}

\end{document}